\documentclass[11pt]{article}

\usepackage[preprint]{acl}

\usepackage{times}
\usepackage{latexsym}

\usepackage[T1]{fontenc}

\usepackage[utf8]{inputenc}

\usepackage{microtype}

\usepackage{inconsolata}

\usepackage{graphicx}
 \usepackage{amsmath}
 \usepackage{enumitem}
\usepackage{float} 
\usepackage{pdfpages}
\usepackage{booktabs}

\title {Multilingual OCR-Aware Fine-Tuning and Prompt-Guided Chain-of-Thought Reasoning for Multimodal Large Language Models}

\author{
Qinwu Xu$^{1}$ \quad Yifan Jiang$^{2}$ \quad Haoyu Ren$^{1,*}$ \\[1ex]
\\
$^{1}$Meta AI \\
$^{2}$UT Austin\\
$^{*}$ Current independent researcher
}

\begin{document}

\maketitle

\begin{abstract}
Optical character recognition (OCR) and multilingual scene-text understanding
remain challenging for multimodal large language models (MLLMs), particularly
in real-world images containing small or degraded text, cluttered layouts,
occlusion, handwriting, and complex typography. We present an OCR-aware
multilingual post-training framework that improves visual-text grounding in a
general-purpose MLLM without requiring an external OCR engine, OCR-extracted
text, or text bounding boxes at inference time. The framework combines
large-scale multilingual OCR supervision, approximately 5M additional
multilingual training samples, controlled synthetic OCR generation and in-image
text translation, LoRA-based supervised fine-tuning (SFT), and lightweight
OCR-oriented Chain-of-Thought prompting. On a held-out real-world multilingual
OCR benchmark, OCR-SFT improves OCR completeness from 71.3 to 84.6, reduces
hallucination rate from 18.3\% to 5.5\%, and improves translation BLEU-1 from
52.3 to 80.2, with substantial hallucination reductions under blur and
rotation. Evaluation on public benchmarks further shows gains on OCR-intensive
tasks while largely preserving broader multimodal capabilities; ablations show
that SFT provides the primary improvement, with prompting offering smaller
complementary gains. These results demonstrate that data-centric OCR-aware
post-training provides a practical and scalable approach to improving
multilingual visual-text grounding in general-purpose MLLMs.
\end{abstract}

\section{Introduction}
\subsection {Motivation}
Text in images is ubiquitous in real-world environments, including menus,
receipts, medication labels, street signs, posters, and documents. Accurate
OCR and multilingual text understanding are therefore important capabilities
for multimodal assistants and vision--language systems. However, recognizing
text and reasoning on the OCR-embedded natural images remains substantially more challenging than conventional document OCR due to blur, low resolution, glare, perspective distortion, partial occlusion, cluttered layouts, and multilingual content.

Recent multimodal large language models (MLLMs), including
Qwen-VL~\cite{bai2023qwenvl}, PaLI-X~\cite{chen2024palix},
LLaVA~\cite{liu2023visual}, Kosmos-2~\cite{peng2023kosmos2},
and Gemini- and GPT-family multimodal systems
~\cite{team2023gemini,openai2023gpt4v}, demonstrate strong visual understanding
and reasoning capabilities. OCR-specialized and document-oriented models have
further advanced visual-text recognition and document understanding
~\cite{hu2024mplugdocowl,wei2024general,chen2025oceanocr}.
Nevertheless, robust OCR grounding in general-purpose MLLMs remains
challenging, particularly under degraded real-world conditions (~\cite{xu2026eval,xu2026grounded}).

In practice, we observe several recurring failure modes: (1) small or degraded
text is skipped or incorrectly recognized; (2) text distributed across multiple
image regions is only partially recovered; and (3) ambiguous or partially
visible text is ``completed'' using language priors rather than grounded visual
evidence. These errors are particularly problematic for general-purpose MLLMs,
where OCR must coexist with broader visual understanding and reasoning rather
than operate as an isolated recognition task.

To address these challenges, we develop an OCR-aware multilingual post-training
framework for general-purpose MLLMs. The framework combines large-scale
multilingual OCR data construction, OCR-aware supervised fine-tuning (SFT),
and lightweight prompt-guided reasoning to improve text grounding under
challenging visual conditions. It further incorporates controlled synthetic OCR
generation and in-image multilingual text translation for scalable data
construction. The resulting model performs OCR and multilingual visual-text
reasoning end-to-end without an external OCR engine at inference time.

\subsection{Related Work}

\paragraph{Multimodal models and visual-text understanding.}
General-purpose MLLMs such as LLaVA~\cite{liu2023visual},
Qwen-VL~\cite{bai2023qwenvl}, AnyMAL~\cite{moon2024anymal},
PaLI-X~\cite{chen2024palix}, Kosmos-2~\cite{peng2023kosmos2},
and InternVL3~\cite{zhu2025internvl3} integrate visual representations with
large language models for joint perception and reasoning. Recent OCR- and
document-oriented models, including mPLUG-DocOwl~1.5
~\cite{hu2024mplugdocowl}, GOT-OCR~2.0~\cite{wei2024general},
and Ocean-OCR~\cite{chen2025oceanocr}, further improve text recognition and
document understanding through specialized architectures, high-resolution
visual encoding, or OCR-focused training.

\paragraph{OCR-aware training and visual grounding.}
Recent approaches improve visual-text understanding through OCR-aware
supervision, synthetic data, and explicit visual grounding. In particular,
SimpleOCR~\cite{peng2026simpleocr} renders textual questions directly into
images during training, encouraging MLLMs to rely on visual evidence rather
than textual shortcuts. These approaches demonstrate the value of stronger
visual-text supervision, but robust multilingual grounding under degraded
real-world conditions remains challenging, particularly when OCR capability
must coexist with general multimodal reasoning.

\subsection{Positioning of Our Work}

Our work is complementary to these directions. We do not introduce a new
multimodal backbone, OCR-specific recognition module, or fine-tuning
algorithm. Instead, we study OCR grounding as a post-training problem for a
general-purpose MLLM. Our contribution is an integrated multilingual OCR
post-training and data framework combining large-scale OCR supervision,
controlled synthetic OCR generation, in-image multilingual text translation,
OCR-aware SFT, and lightweight grounding-oriented prompting. A key
system-design distinction is that OCR capability is integrated directly into
the MLLM through post-training rather than supplied by an auxiliary OCR
pipeline: no external OCR engine, OCR-extracted text, or text bounding boxes
are required at inference time.

Our contributions are summarized as follows:
\begin{enumerate}[label=(\roman*),leftmargin=*, labelsep=0.5em, nosep]

\item We introduce an OCR-aware multilingual post-training framework for
improving OCR grounding and multilingual visual-text understanding in a
general-purpose MLLM while preserving broader multimodal capabilities.

\item We construct a large-scale multilingual OCR corpus covering diverse
real-world content and challenging visual conditions, including blur, occlusion,
low resolution, rotation, dense layouts, and multilingual scene text.

\item We develop complementary data-construction mechanisms for scalable
synthetic OCR generation and in-image multilingual text translation, enabling
controlled text modification while preserving scene context and visual
consistency.

\item Experiments on held-out real-world data and public benchmarks demonstrate
improved OCR completeness, reduced hallucination, and stronger text-grounded
understanding while maintaining general visual reasoning performance. We
publicly release a subset of our real-world benchmark, including particularly
challenging OCR cases, together with the evaluation pipeline to support
reproducibility.\footnote{\url{https://huggingface.co/datasets/vlmgrounding/ocr_ground}}

\end{enumerate}

\section{Methodologies}

\subsection{Problem Formulation}

We study multimodal generation under noisy OCR conditions, where textual
information is embedded in an image and must be inferred jointly with other
visual evidence. Let $I$ denote an input image, $q$ a user query, and $Y$ the
target response, such as recognized text, an answer, or a translation. We
model the task directly as
\begin{equation}
    p_{\theta}(Y \mid I,q),
\end{equation}
without explicitly detecting text regions or invoking an external OCR engine
at inference time.

Under challenging conditions such as blur, occlusion, low resolution,
perspective distortion, and cluttered layouts, visual evidence for the
underlying text can be incomplete or ambiguous. Autoregressive MLLMs may then
rely excessively on learned language priors, resulting in omitted,
incorrectly recognized, or visually unsupported text.

We therefore optimize the model end-to-end using supervised fine-tuning (SFT):
\begin{equation}
    \mathcal{L}_{\mathrm{SFT}}
    = - \sum_{t=1}^{|Y|}
    \log p_{\theta}(y_t \mid y_{<t}, I, q).
    \label{eq:sft}
\end{equation}
The primary objective is standard autoregressive cross-entropy; our contribution
lies in the OCR-aware multilingual supervision and post-training framework
described below, which are designed to improve grounding under challenging
visual conditions.

\subsection{Model Architecture}

\begin{figure} [t]
    \centering
    \includegraphics[width=0.75\linewidth]{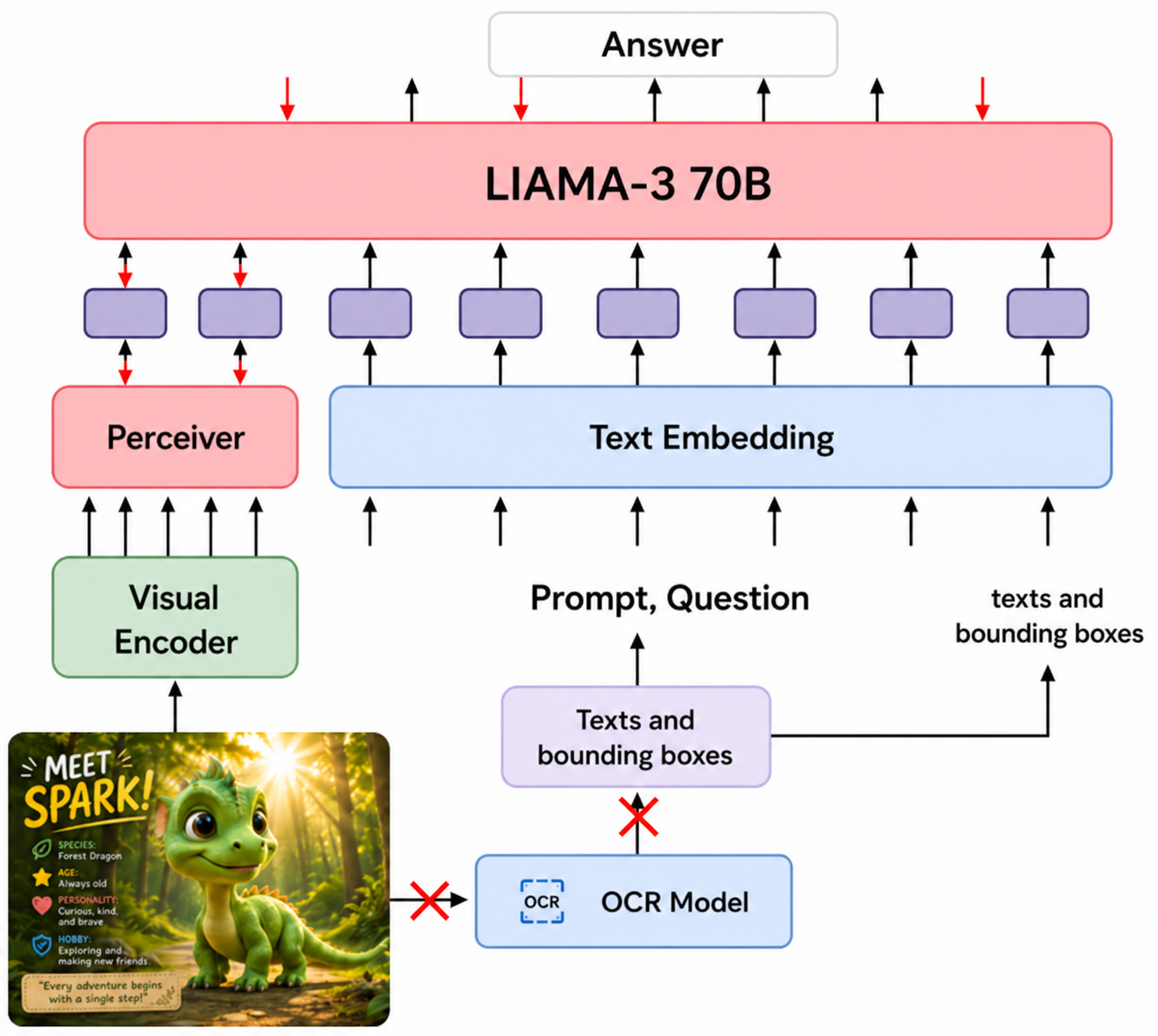}
    \caption{Model architecture}
    \label{fig:architecture2}
\end{figure}

We use a general-purpose MLLM architecture following the broad design of
large-scale multimodal systems such as AnyMAL~\cite{moon2024anymal}, with
LLaMA-3 70B as the language backbone. As illustrated in
Figure~\ref{fig:architecture2}, a frozen MetaCLIP-based ViT
encoder~\cite{xu2023metaclip} extracts visual features, which are compressed
into a fixed number of visual tokens by a Perceiver-based
resampler~\cite{jaegle2021perceiver}. The resulting visual tokens are
projected into the language embedding space and processed together with text
tokens by the LLaMA decoder for end-to-end multimodal generation.

Our work does not introduce a new multimodal backbone or OCR-specific
recognition module. Instead, a key system-design contribution is to integrate
OCR capability directly into the general-purpose MLLM through OCR-aware
post-training, rather than relying on an auxiliary OCR pipeline at inference.
Unlike conventional OCR pipelines that separate text detection and recognition
~\cite{zhou2017east,baek2019craft,smith2007tesseract,li2021trocr}, or systems
that provide OCR-extracted text or detected regions as additional inputs, our
model receives only the image and user query at inference time. No external
OCR model, OCR-extracted text, or text bounding boxes are required. OCR
recognition and downstream language reasoning are therefore performed jointly
within the end-to-end multimodal model.

\subsection{OCR Data Curation}

We curate a large-scale multilingual OCR corpus designed to capture the
diversity and degradation encountered in real-world visual-text understanding.
The corpus primarily covers five languages: English, Spanish, French, Italian, and German. Starting from approximately 1M source OCR examples, we construct approximately 5M additional multilingual samples by generating about five target-language variants per source example on average. These OCR data are combined with the general multimodal training data during post-training to improve text grounding while preserving broader visual capabilities.

The corpus covers diverse visual domains. Its approximate composition is:
Text/Chart/Document (23.7\%), Postcards/News (5.6\%), Gardening/Plants
(8.3\%), Landmarks (13.2\%), Foods (13.3\%), Shopping (15.3\%), Indoor
Activity (10.8\%), and Outdoor Activity (9.8\%). It includes receipts, menus,
brochures, advertisements, handwritten notes, street signs, retail products,
documents, educational materials, maps, charts, and mixed text-image scenes.
We further include challenging conditions such as dense layouts, handwriting,
rotation, low contrast, blur, occlusion, and degraded image quality.

Beyond direct text recognition, the corpus contains image--question--answer
supervision requiring OCR-grounded reasoning. Tasks include text extraction,
multilingual translation, information retrieval, document comprehension,
summarization, entity extraction, and computation over visual-text evidence, etc.

\subsubsection{Synthetic Multilingual OCR Data Generation}

\begin{figure}[t]
\centering
\includegraphics[width=\linewidth]{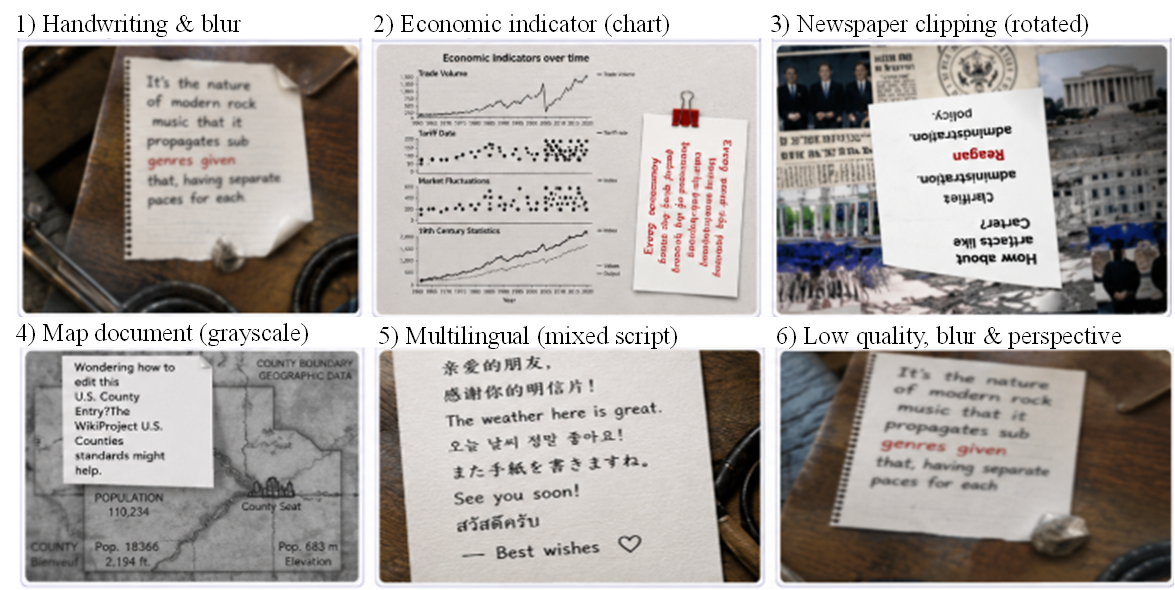}
\caption{Representative synthetic multilingual OCR examples with diverse
visual conditions, including blur, rotation, handwriting, and complex layouts.}
\label{fig:diffusion-image2}
\end{figure}

To scale multilingual coverage while controlling the underlying textual
content, we construct synthetic OCR-grounded examples by translating source
text into target languages and rendering the translated content into
corresponding visual scenes. We additionally introduce OCR-relevant
degradations such as blur, rotation, occlusion, perspective distortion,
compression artifacts, and cluttered layouts. Representative examples are
shown in Figure~\ref{fig:diffusion-image2}.

Unlike unconstrained text-to-image generation, this construction explicitly
controls the target OCR content and its visual placement, providing known
textual supervision while allowing variation in scene appearance and image
quality. The resulting examples complement real-world data with controlled
multilingual and degraded OCR conditions.

\subsubsection{Modular In-Situ Visual Translation}
\label{sssec:modular_translation_framework}

\begin{figure}[t]
\centering
\includegraphics[width=0.75\linewidth]{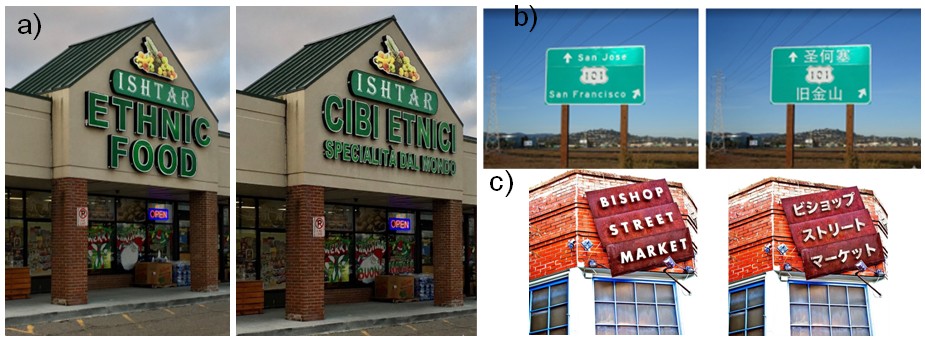}
\caption{Examples of in-situ visual text translation from English to
(a) Spanish (image source: Andre Carrotflower, CC BY-SA 4.0),
(b) Chinese (photo captured by the authors), and
(c) Japanese (image source: Drumguy8800, CC BY-SA 3.0).}
\label{fig:img_translate}
\end{figure}

As a complementary data-construction strategy, we develop a modular framework
for replacing text within existing images while preserving the surrounding
visual context (Figure~\ref{fig:img_translate}). The framework consists of
three stages: (i) \textit{text localization}, using SAM~2 to obtain pixel-level
masks for source-text regions; (ii) \textit{background reconstruction}, using
generative inpainting to remove the source text and reconstruct the local
background; and (iii) \textit{style-adaptive rendering}, which estimates
visual attributes such as typography, color, and local appearance and renders
the translated text with geometric alignment and alpha blending.

This framework is used exclusively for offline training-data construction.
SAM~2, inpainting, OCR-extracted text, and other image-editing components are
not used by the MLLM at inference time. Compared with end-to-end generative
image translation, the modular construction separates text replacement from
background reconstruction, providing direct control over translated content,
placement, and local appearance while better preserving the original scene.
Additional implementation details and comparisons are provided in
Appendix~A and Table~\ref{tab:diffusion-modular}.

\subsection{Post-Training Framework}

\subsubsection{OCR-Aware Supervised Fine-Tuning}

We perform OCR-aware supervised fine-tuning (SFT) on the multilingual
image--question--answer corpus described in Section~\ref{sec:ocr_data}.
The training data combine general multimodal supervision with approximately
5M additional OCR-oriented multilingual samples, allowing the model to
improve visual-text grounding while retaining broader multimodal capabilities.
The OCR data cover text recognition, translation, and OCR-grounded reasoning
under both natural and synthetically degraded visual conditions.

We use LoRA for parameter-efficient adaptation of the LLaMA-3 70B backbone,
with rank $r=256$ and scaling parameter $\alpha=512$, while keeping the
pretrained backbone weights frozen. Training uses AdamW with a peak learning
rate of $1\times10^{-5}$, linear warm-up followed by cosine decay, a global
batch size of 64, gradient accumulation of 1, BF16 precision, and one training
epoch. The standard autoregressive SFT objective in Eq.~\ref{eq:sft} is used
throughout training.

Training is distributed over 256 NVIDIA H100 80GB GPUs using DeepSpeed
ZeRO-3 and FlashAttention, with high-bandwidth InfiniBand communication
across nodes. One epoch over approximately 5M OCR-oriented training instances
requires about 30 hours.

\subsubsection{OCR-Oriented Chain-of-Thought Reasoning}

OCR-aware SFT provides the primary source of improvement, but challenging
cases may still contain ambiguous, partially visible, or degraded text. As an
auxiliary mechanism, we incorporate OCR-oriented Chain-of-Thought (CoT)
instructions during post-training to encourage the model to inspect relevant
visual evidence before producing an answer.

Let $R$ denote an intermediate reasoning trace induced by the prompt.
Conceptually, generation can be expressed as
\begin{equation}
R = \mathrm{CoT}(I,q), \qquad
Y \sim p_{\theta}(Y \mid I,q,R).
\end{equation}
Here, $R$ is not separately supervised; it is induced through structured
OCR-aware instructions that encourage attention to relevant image regions,
uncertain text, and observable visual evidence. This prompting is intended to
reduce reliance on language priors when OCR evidence is incomplete or
ambiguous.

Ablations in Section~\ref{sec:results} show that OCR-oriented CoT provides
modest complementary gains beyond OCR-aware SFT, particularly for challenging
visual-text reasoning cases. Prompting details are provided in Appendix~B.

\subsection{Evaluation Protocol and Metrics}

We evaluate OCR capability using three primary metrics:

\begin{enumerate}[label=(\roman*), itemsep=0pt, topsep=0pt]

\item \textbf{OCR Completeness:} token-level recall between the normalized
reference OCR text and model output, measuring the fraction of reference
content successfully recovered. Text is normalized before matching to reduce
sensitivity to formatting and superficial tokenization differences.

\item \textbf{Hallucination Rate:} percentage of responses containing OCR
content that is unsupported by the visible text or inconsistent with the
reference evidence. Responses are evaluated using an LLM-based judge, with
low-confidence cases (confidence $<0.85$) manually verified.

\item \textbf{Translation Quality:} BLEU-1 between the generated and reference
translations. We use unigram BLEU to emphasize lexical fidelity of recognized
and translated OCR content, while completeness and hallucination are evaluated
separately by the preceding metrics.

\end{enumerate}

Our primary evaluation uses approximately 3,000 held-out image--question--answer
samples that are excluded from post-training. The benchmark follows the mixed
multilingual distribution encountered in deployment and primarily covers
English, Spanish, French, Italian, and German. It contains OCR
understanding and translation tasks across diverse real-world scenarios,
including documents, menus, signs, handwritten content, retail products, and
indoor and outdoor scenes. Reference OCR text and answers are constructed from
available annotations and OCR-assisted labeling, with human validation for
ambiguous or uncertain cases.

We compare the OCR-aware model with its pre-post-training LLaMA-3 VLM
baseline to isolate the effect of OCR-aware adaptation. To evaluate
generalization beyond the internal benchmark, we additionally report results
on public benchmarks including DocVQA, TextVQA, MMBench, and VQAv2, with
Qwen2.5-VL-7B and InternVL3-8B as external baselines. Qualitative comparisons
with GPT- and Gemini-family models are used only to illustrate representative
failure modes and challenging real-world cases.

\section{Results}
\label{sec:results}

\subsection{Quantitative Results}

On the held-out real-world OCR benchmark, OCR-aware SFT substantially improves
all three OCR-specific metrics (Table~\ref{tab:ocr-translation-accuracy}).
OCR completeness increases from 71.3 to 84.6, while hallucination rate
decreases from 18.3\% to 5.5\%. Translation BLEU-1 also increases from 52.3
to 80.2. These results show consistent improvements in text recovery,
grounding, and multilingual translation after OCR-aware post-training.

\begin{table}[t]
\centering
\caption{Results on the held-out real-world multilingual OCR benchmark.
Higher is better for completeness and BLEU-1; lower is better for
hallucination rate.}
\label{tab:ocr-translation-accuracy}
\resizebox{\columnwidth}{!}{
\begin{tabular}{lccc}
\toprule
Model & Completeness $\uparrow$ & Hallucination $\downarrow$ &
BLEU-1 $\uparrow$ \\
\midrule
Baseline VLM & 71.3 & 18.3 & 52.3 \\
OCR-SFT      & \textbf{84.6} & \textbf{5.5} & \textbf{80.2} \\
\bottomrule
\end{tabular}}
\end{table}

The reduction in hallucination persists under visual degradation
(Table~\ref{tab:hallucination-rates}). Compared with the baseline, OCR-SFT
reduces hallucination from 8.0\% to 2.2\% on clean images, from 24.8\% to
6.6\% under blur, and from 21.4\% to 5.8\% under rotation. The larger
absolute improvements under degraded conditions suggest that OCR-aware
training is particularly useful when visual-text evidence becomes unreliable.

\begin{table}[t]
\centering
\caption{Hallucination rate (\%, $\downarrow$) under different image
conditions.}
\label{tab:hallucination-rates}
\resizebox{0.75\columnwidth}{!} {
\begin{tabular}{lcc}
\toprule
Condition & Baseline VLM & OCR-SFT \\
\midrule
Clean    & 8.0  & \textbf{2.2} \\
Blur     & 24.8 & \textbf{6.6} \\
Rotation & 21.4 & \textbf{5.8} \\
\bottomrule
\end{tabular}
}
\end{table}

\subsection{Public Benchmarks and Ablation}

To evaluate generalization beyond our held-out OCR benchmark, we further
compare against Qwen2.5-VL-7B and InternVL3-8B on standard public benchmarks.
We also ablate OCR-SFT and OCR-oriented CoT to separate their contributions.

\begin{table}[t]
\centering
\caption{Public benchmark results and post-training ablation. Best results
are bolded.}
\label{tab:public-benchmarks}
\resizebox{\columnwidth}{!}{
\begin{tabular}{lcccc}
\toprule
Model & DocVQA & TextVQA & MMBench & VQAv2 \\
      & EM-Norm $\uparrow$ & VQA-Soft $\uparrow$ &
        MCQ $\uparrow$ & VQA-Soft $\uparrow$ \\
\midrule
Qwen2.5-VL-7B & \textbf{87.08} & 72.77 & \textbf{82.60} & 67.81 \\
InternVL3-8B   & 75.02 & 63.35 & 82.49 & 61.86 \\
\midrule
Baseline VLM   & 81.5 & 80.5 & 78.0 & \textbf{81.5} \\
+ OCR-SFT      & 82.3 & 82.0 & 79.0 & 81.4 \\
+ OCR-SFT+CoT  & 82.5 & \textbf{82.2} & 78.9 & 81.4 \\
\bottomrule
\end{tabular}}
\end{table}

OCR-SFT improves the baseline on the OCR-intensive DocVQA and TextVQA
benchmarks, from 81.5 to 82.3 and from 80.5 to 82.0, respectively.
OCR-oriented CoT provides a smaller additional improvement, reaching 82.5
and 82.2. In contrast, VQAv2 remains essentially unchanged (81.5 vs. 81.4),
while MMBench improves slightly after OCR-SFT (78.0 to 79.0). These results
suggest that the gains are primarily attributable to OCR-aware SFT and that
OCR specialization does not substantially degrade broader visual reasoning
performance.

Compared with public MLLMs, our OCR-post-trained model does not dominate
every benchmark: Qwen2.5-VL-7B achieves higher DocVQA and MMBench scores.
However, the proposed model performs substantially better on TextVQA and
VQAv2 in this evaluation. Together with the within-model ablation, these
results support the intended objective of improving OCR grounding while
preserving general multimodal capability.

\subsection{Qualitative Analysis and Inference Examples}

We further examine model behavior on representative in-the-wild OCR examples
covering multilingual text, dense layouts, partially visible text, embedded
scene text, and OCR-dependent visual reasoning. These examples complement the
quantitative evaluation by illustrating common failure modes and behavioral
differences before and after OCR-aware post-training.

\begin{figure}[t]
\centering
\includegraphics[width=\linewidth]{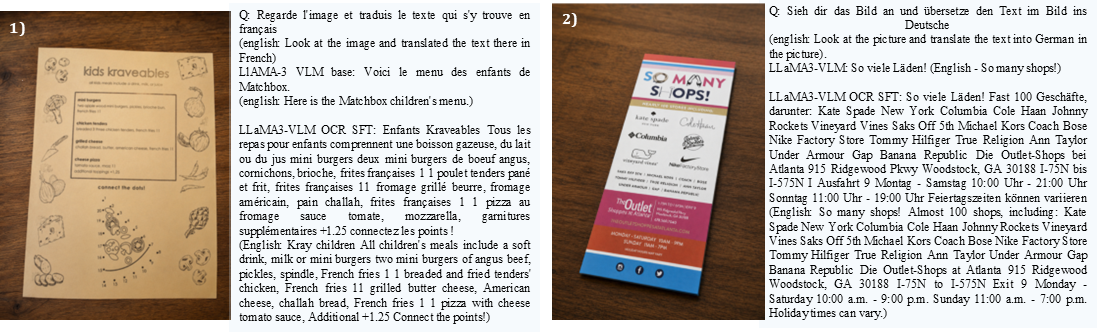}
\caption{Representative multilingual OCR translation examples comparing the
baseline LLaMA3-VLM and OCR-SFT model. OCR-SFT recovers more textual content
and produces more complete French and German translations under sparse and
dense OCR conditions. The left image is AI-generated; the right image is a
brochure from the Outlet Shoppes at Atlanta.}
\label{fig:menu_broucher}
\end{figure}

Figure~\ref{fig:menu_broucher} compares the baseline and OCR-SFT models on
multilingual OCR translation under sparse and dense text conditions. The
baseline often produces coarse scene-level descriptions (e.g., identifying a
``menu'' or ``receipt'') while omitting detailed textual content. In the shown
examples, OCR-SFT recovers more fine-grained information, including menu items,
prices, structured fields, and their multilingual translations, particularly
in dense layouts containing small or visually cluttered text.

\begin{figure}[t]
\centering
\includegraphics[width=\linewidth]{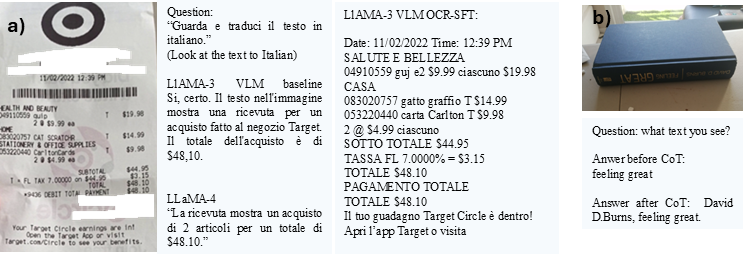}
\caption{Representative OCR grounding examples. (a) Extraction and translation
of dense receipt content. (b) Recognition of distant, blurred scene text with
OCR-oriented prompt-guided reasoning.}
\label{fig:receipt}
\end{figure}

Figure~\ref{fig:receipt} illustrates two additional challenging cases.
In Example~(a), OCR-SFT recovers dense itemized receipt information, including
fields and prices, that is partially missed by the baseline. Example~(b)
illustrates the auxiliary OCR-oriented CoT strategy on distant, blurred scene
text. With the structured reasoning prompt, the model inspects the relevant
visual evidence before answering and recovers the text ``David D. Burns,''
which is omitted under direct generation. This example illustrates how
prompt-guided reasoning can help in difficult cases, while the quantitative
ablation in Table~\ref{tab:public-benchmarks} shows that its aggregate gain
over OCR-SFT is modest.

\begin{figure}[t]
\centering
\includegraphics[width=\linewidth]{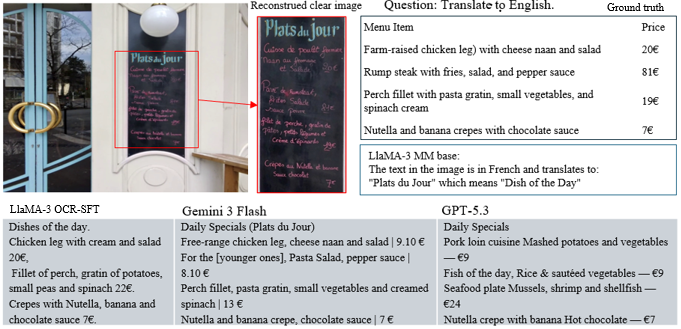}
\caption{Representative cross-model comparison on handwritten French scene
text with distortion, variable formatting, and a complex background.}
\label{fig:menu_wall}
\end{figure}

We additionally present qualitative comparisons with GPT- and Gemini-family
models to illustrate challenging cross-model failure cases rather than provide
a systematic ranking of frontier systems. In Figure~\ref{fig:menu_wall}, which
contains handwritten French text under difficult visual conditions, our
OCR-SFT model correctly identifies three menu items and two corresponding
prices. Gemini recovers more items but only one accurate price, while GPT
recovers fewer correct items and prices. The example highlights that extraction
completeness and grounding accuracy can differ substantially even among strong
multimodal models.

Overall, these examples are consistent with the quantitative results:
OCR-aware SFT improves recovery of difficult visual text and reduces unsupported
generation, while OCR-oriented CoT can provide additional assistance in some
ambiguous cases. The examples also illustrate a qualitative shift from coarse
scene-level responses toward more detailed, visually grounded text extraction
and reasoning after OCR-aware post-training.

\subsection{Deployment Efficiency and Practical Considerations}

A practical motivation for integrating OCR capability directly into the MLLM
is to avoid a separate OCR inference pipeline. We therefore measure end-to-end
inference efficiency under a representative serving configuration using
4$\times$ NVIDIA A100-SXM4-80GB GPUs. As shown in
Table~\ref{tab:deployment}, processing 10 representative multimodal samples
requires 21.57\,s, corresponding to a mean latency of 2.16\,s per sample.
The measured P50 and P90 latencies are 2.03\,s and 2.74\,s, respectively,
with an end-to-end throughput of 0.464 samples/s.

\begin{table}[t]
\centering
\caption{End-to-end inference efficiency on 4$\times$ A100-SXM4-80GB GPUs.}
\label{tab:deployment}
\begin{tabular}{lc}
\toprule
Metric & Measurement \\
\midrule
Total time (10 samples) & 21.57 s \\
Mean latency & 2.16 s/sample \\
P50 latency & 2.03 s/sample \\
P90 latency & 2.74 s/sample \\
Throughput & 0.464 samples/s \\
\bottomrule
\end{tabular}
\end{table}

These measurements include the complete multimodal inference path rather than
an isolated OCR component. Because OCR recognition is learned within the MLLM,
deployment does not require an additional text detector, OCR recognizer, or
OCR-to-LLM communication stage. This simplifies the serving pipeline and
avoids additional OCR-specific inference components, although the overall
latency remains dominated by generation with the 70B language backbone.


\section{Conclusion}

We present an OCR-aware multilingual post-training framework for improving
visual-text grounding in general-purpose MLLMs under challenging real-world
conditions. The framework combines large-scale multilingual OCR supervision,
controlled synthetic data generation, in-image visual text translation,
OCR-aware SFT, and lightweight prompt-guided reasoning, while requiring no
external OCR engine or text detector at inference time.

Experiments on held-out real-world data show substantial improvements in OCR
completeness, multilingual translation, and hallucination rate, including under
degraded visual conditions such as blur and rotation. Evaluation on public
benchmarks further shows improvements on OCR-intensive tasks while largely
preserving broader multimodal capabilities. OCR-aware SFT provides the primary
gain, with prompt-guided reasoning offering smaller complementary improvements
on challenging cases.

Overall, the results demonstrate that targeted OCR-aware post-training provides
a practical approach for improving multilingual text grounding in
general-purpose MLLMs without introducing an OCR-specific inference pipeline.

\section{Limitations}

Our evaluation primarily targets multilingual OCR and text-grounded visual
understanding, with the largest-scale experiments focused on six languages.
Although public benchmarks provide additional evidence of generalization,
broader evaluation across languages, domains, and multimodal reasoning tasks
would further establish the scope of the observed gains. In addition, synthetic
OCR data may not fully capture rare real-world degradations and typography,
and highly ambiguous, occluded, or low-resolution text can still lead to
incorrect or unsupported generations. Finally, the reported deployment
measurements reflect our evaluated hardware configuration and may vary with
model serving infrastructure and optimization.

\bibliography{custom}

\section*{Appendix A: Detailed OCR Data Generation Pipeline}

\subsection*{Synthetic Multilingual OCR Data Generation}

We construct synthetic OCR-grounded data to jointly control multilingual
textual content and its visual realization. Let $T_{\mathrm{src}}$ denote
source OCR text and $T_{\mathrm{tgt}}$ its translation into a target language.
The generation process can be represented as
\begin{equation}
\small
T_{\mathrm{src}} \sim p(T), \qquad
T_{\mathrm{tgt}} \sim
p(T_{\mathrm{tgt}} \mid T_{\mathrm{src}}), \qquad
I \sim p(I \mid T_{\mathrm{src}}, T_{\mathrm{tgt}}),
\end{equation}
which induces the joint distribution
\begin{equation}
\small
p(I,T_{\mathrm{src}},T_{\mathrm{tgt}})
=
p(T_{\mathrm{src}})
p(T_{\mathrm{tgt}} \mid T_{\mathrm{src}})
p(I \mid T_{\mathrm{src}},T_{\mathrm{tgt}}).
\end{equation}

This construction couples visual inputs with known multilingual textual
supervision. Rather than relying on unconstrained text-to-image generation,
the pipeline explicitly controls the target OCR content and its placement
while allowing variation in typography, layout, scene appearance, and image
quality. This provides scalable supervision for text recognition, translation,
and OCR-grounded multimodal reasoning.

The synthetic data complement the real-world OCR corpus by increasing
coverage of multilingual and long-tail visual-text conditions. Visual
degradations and other OCR perturbations are applied as described in the main
text, allowing the same underlying textual supervision to be presented under
different levels of visual uncertainty.

\subsection*{Modular In-Situ Visual Translation Framework}

As a complementary construction strategy, we use a modular framework to
replace source text within existing images while preserving the surrounding
scene. The pipeline separates localization, background reconstruction, and
translated-text rendering into three stages:

\begin{enumerate}[label=(\roman*), leftmargin=0pt, itemindent=1.5em]

\item \textit{Text localization:}
Text regions are localized using DBNet or SAM~2 to obtain masks identifying
the source-text regions to be modified.

\item \textit{Background reconstruction:}
After removal of the source text, the corresponding regions are reconstructed
using generative inpainting while retaining the surrounding non-textual scene
content.

\item \textit{Style-adaptive rendering:}
Translated text is rendered into the reconstructed regions using geometric
transformation and alpha blending. Local appearance information, including
text geometry, color, and typographic characteristics when recoverable, is
used to better match the translated text to the original scene.

\end{enumerate}

Separating these operations provides direct control over the translated text,
its spatial placement, and the region being modified, rather than requiring a
generative model to jointly regenerate the entire image and textual content.
Importantly, DBNet, SAM~2, inpainting, and rendering are used only for offline
training-data construction. None of these components, nor OCR-extracted text
or bounding boxes, are required by the fine-tuned MLLM at inference time.

\subsection*{Comparison with End-to-End Diffusion Approaches}

End-to-end diffusion-based image generation jointly synthesizes scene content
and text appearance, providing substantial flexibility for open-ended image
generation but less explicit control over exact textual content. For OCR
training-data construction, generation errors in characters, typography, or
text placement can directly corrupt the intended supervision.

Our modular framework instead separates background reconstruction from text
replacement. This provides explicit control over the translated string and
its placement while preserving as much of the original scene as possible.
The two approaches therefore serve different objectives: end-to-end
generation emphasizes flexible scene synthesis, whereas our modular pipeline
emphasizes controllable modification of text within an existing visual
context. Table~\ref{tab:diffusion-modular} summarizes these differences.

\begin{table}[t]
\caption{Comparison of end-to-end (E2E) diffusion-based generation and the
modular in-situ visual translation framework.}
\label{tab:diffusion-modular}
\centering
\resizebox{\columnwidth}{!}{
\begin{tabular}{
p{2.3cm}
@{\hspace{0.15cm}}
p{4.2cm}
@{\hspace{0.15cm}}
p{4.2cm}
}
\toprule
Requirement & E2E Diffusion & Modular Framework \\
\midrule
Primary objective
& Open-ended image synthesis
& In-situ text transformation \\

Background treatment
& Jointly generated with scene
& Explicitly reconstructed and preserved \\

Text control
& Generative; susceptible to text artifacts
& Explicitly controlled during rendering \\

Design focus
& Global semantic and visual coherence
& Text fidelity and local visual consistency \\

Typical use
& General image generation
& Structured multilingual OCR data construction \\
\bottomrule
\end{tabular}
}
\end{table}

\section*{Appendix B: OCR-Oriented Chain-of-Thought Prompting}

The OCR-oriented prompt encourages evidence-grounded reasoning without
requiring separately annotated reasoning traces. For challenging OCR queries,
the model is instructed to: (1) first consider the overall scene and context;
(2) inspect relevant image regions, particularly when text is small, distant,
partially visible, or spatially distributed; (3) reason cautiously about text
affected by blur, rotation, low contrast, or occlusion; and (4) base the final
response on observable visual evidence rather than completing uncertain text
from language priors.

For simple OCR queries, the model may answer directly; explicit multi-step
reasoning is not required for every input. The prompting mechanism is therefore
used as an auxiliary grounding strategy rather than as a requirement to
generate long reasoning traces.

\paragraph{Prompt template.}
The system instruction asks the model to first inspect the overall scene,
identify visually relevant text regions, reason cautiously when text is
degraded or partially visible, and base the final response only on observable
visual evidence. For straightforward OCR queries, the model is allowed to
answer directly without producing an explicit reasoning trace.

\section*{Appendix C: Qualitative Comparison Across Models}

We present 15 representative challenging examples selected from a broader
qualitative analysis of more than 100 OCR cases spanning multilingual
documents, receipts, menus, brochures, handwritten text, scene text, and
degraded OCR conditions. These examples are intended to illustrate recurring
failure modes rather than provide a systematic ranking of the compared models.

Across the selected cases, several behavioral patterns are observed. Gemini
often produces detailed responses but can introduce visually unsupported
details when OCR evidence is ambiguous. GPT tends to produce more conservative
responses but can omit content or infer plausible text under noisy OCR
conditions. The OCR-SFT model often produces less expansive responses while
remaining closely grounded in visible text, particularly in several
handwritten, low-contrast, and degraded examples. All models degrade under
severe blur, occlusion, mirrored text, and other ambiguous conditions.

Safety-related refusal behavior also differs across systems; because such
behavior reflects model-specific alignment policies in addition to visual-text
capability, we do not interpret these cases as direct measures of OCR quality.
Table~\ref{tab:model_comparison} summarizes the qualitative patterns observed
in the selected examples, with the complete examples provided on the following
pages.

\begin{table*}[!t]
\centering
\caption{Qualitative behavioral patterns observed in the selected OCR-heavy
examples. These observations are illustrative rather than a systematic model
ranking.}
\label{tab:model_comparison}
\footnotesize
\setlength{\tabcolsep}{6pt}
\renewcommand{\arraystretch}{1.1}

\begin{tabular*}{\textwidth}{@{\extracolsep{\fill}}p{3cm}p{6.5cm}p{6.5cm}}
\toprule
\textbf{Model} & \textbf{Observed strengths} & \textbf{Observed failure modes} \\
\midrule
Gemini 3 Flash &
Detailed extraction and structured responses &
Over-generation and unsupported details in some ambiguous cases \\

LLaMA-3 VLM OCR-SFT &
Strong OCR grounding with fewer unsupported details &
Sometimes less detailed than other models \\

GPT-5.3 (instant variant) &
Generally conservative responses &
Missing or inferred details in some ambiguous OCR cases \\
\bottomrule
\end{tabular*}
\end{table*}

\clearpage

\includepdf[pages=-,landscape=true]{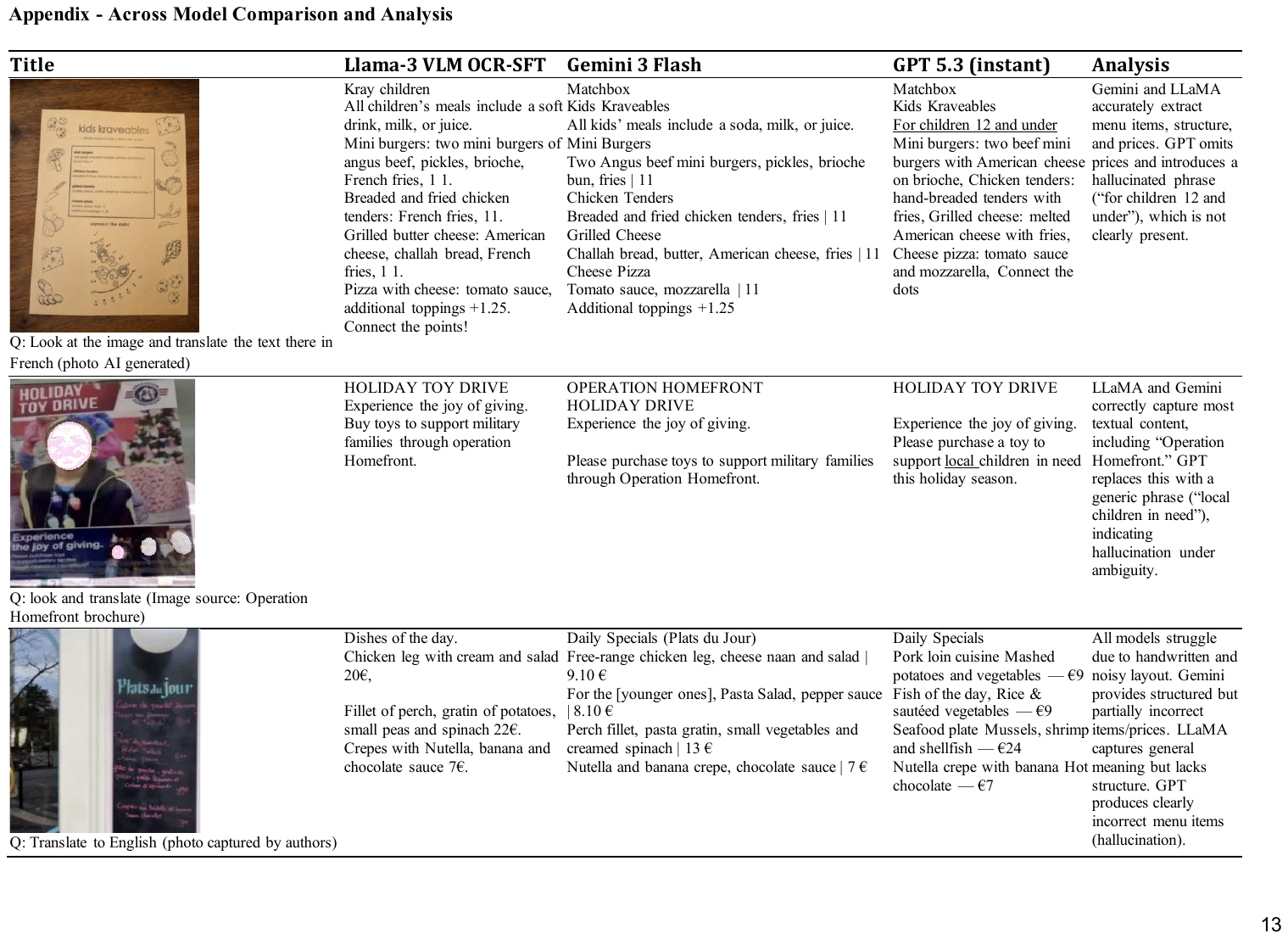}
 
\end{document}